\documentclass[11pt]{article}
\usepackage{coling2020}
\usepackage{times}
\usepackage{url}
\usepackage{latexsym}

\usepackage{algorithmic}
\usepackage{graphicx}
\usepackage{textcomp}
\usepackage{xcolor}
\usepackage{subcaption}
\usepackage{mathtools}
\usepackage{adjustbox}
\usepackage{multirow}
\usepackage{dblfloatfix}
\usepackage{tabularx}
\usepackage{tablefootnote}
\usepackage{makecell}
\usepackage{microtype}
\usepackage{colortbl}
\usepackage{soul}
\usepackage{color}
\usepackage{enumitem}
\usepackage{amssymb}
\newcolumntype{b}{>{\hsize=1.5\hsize}X}
\newcolumntype{s}{>{\hsize=.5\hsize}X}

\colingfinalcopy % Uncomment this line for the final submission

\title{Detect All Abuse! Toward Universal Abusive Language Detection Models}

\author{Kunze Wang\textsuperscript{1*}, Dong Lu\textsuperscript{1*}, Soyeon Caren Han\textsuperscript{2**}, Siqu Long\textsuperscript{1}, Josiah Poon\textsuperscript{2} \\
  The University of Sydney, Sydney, Australia \\
  {\tt \textsuperscript{1}\{kwan4418,dolu4031,slon6753\}@uni.sydney.edu.au}\\  
  {\tt \textsuperscript{2}\{caren.han,josiah.poon\}@sydney.edu.au}  }

\date{}

\begin{document}
\maketitle
\abovedisplayskip=0pt
\abovedisplayshortskip=0pt
\belowdisplayskip=0pt
\belowdisplayshortskip=0pt
\abovecaptionskip=0pt
\belowcaptionskip=0pt
\begin{abstract}
Online abusive language detection (ALD) has become a societal issue of increasing importance in recent years. Several previous works in online ALD focused on solving a single abusive language problem in a single domain, like Twitter, and have not been successfully transferable to the general ALD task or domain. In this paper, we introduce a new generic ALD framework, MACAS, which is capable of addressing several types of ALD tasks across different domains. Our generic framework covers multi-aspect abusive language embeddings that represent the target and content aspects of abusive language and applies a textual graph embedding that analyses the user's linguistic behaviour. Then, we propose and use the cross-attention gate flow mechanism to embrace multiple aspects of abusive language. Quantitative and qualitative evaluation results show that our ALD algorithm rivals or exceeds the six state-of-the-art ALD algorithms across seven ALD datasets covering multiple aspects of abusive language and different online community domains. The code can be downloaded from https://github.com/usydnlp/MACAS.
\end{abstract}
\blfootnote{* Equal contribution }
\blfootnote{** Corresponding author (Caren.Han@sydney.edu.au)}
\section{Introduction}
Abusive language in online communities has become a significant societal problem \cite{nobata2016abusive} and online abusive language detection (ALD) aims to identify any type of insult, vulgarity, or profanity that debases a target or group online. It is not only limited to detecting offensive language \cite{razavi2010offensive}, cyberbullying \cite{xu2012learning}, and hate speech \cite{djuric2015hate}, but also to more nebulous or implicit forms of abuse. Many social media companies and researchers have utilised multiple resources, including machine learning, human reviewers and lexicon-based text analytics to detect abusive language \cite{waseem2016you,qian2018leveraging}. However, none of them can perfectly resolve the ALD task because of the difficulties of moderating user content and in classifying ambiguous posts \cite{CadeMetz2019Facebook}. On the technical side, previous ALD models were developed on only a few subtasks (e.g. hate speech, racism, sexism) in a single domain (like Twitter), and each specialised model is not successfully transferable to general ALD in different online communities.

Our research question is, ``What would be the best generic ALD model that can be used for different types of abusive language detection sub-tasks and in different online communities?" 
To solve this, we found that \newcite{waseem2017understanding} reviewed the existing online abusive language detection literature, and defined a generic abusive language typology that can encompass the targets of a wide range of abusive language subtasks in different types of domain. The typology is categorised in the following two aspects: \textbf{1) Target aspect}: The abuse can be directed towards either a) a specific individual/entity or b) a generalised group. This is an essential sociological distinction as the latter refers to a whole category of people, like a race or gender, rather than a specific individual or organisation; \textbf{2) Content aspect}: The abusive content can be explicit or implicit. Whether directed or generalised, explicit abuse is unambiguous in its potential to be damaging, while implicit abusive language does not immediately imply abuse (through the use of sarcasm, for example). For example, assume that we have a tweet ``F***''. ``You are sooo sweet like other girls''. It includes all those aspects; the directed target (``yourself''), the generalised target (``girls''), the explicit content (``F***''), and the implicit content (``You are sooo sweet''). 

Inspired by this abusive language typology, we propose a new generic ALD framework, MACAS (\textbf{M}ulti-\textbf{A}spect \textbf{C}ross \textbf{A}ttention \textbf{S}uper Joint for ALD), using aspect models and a cross-attention aspect gate flow. First, we build four different types of abusive language aspect embeddings, including directed target, generalised target, explicit content, and implicit content. We also propose to use a heterogeneous graph to analyse the linguistic behaviour of each author and learn word and document embeddings with graph convolutional networks (GCNs). Not every online community (e.g. news forums) allows user-to-user relationship (e.g. follower-following), so we avoid using user-community relationship information. Then, we propose a cross-attention aspect gate flow to obtain the mutual enhancement between the two  aspects. The gate flow contains two gates, target gate and content gate, then fuses the outputs of those gates. The target gate draws on the content probability distribution, utilising the semantic information of the whole input sequence along with the target source, while the content gate takes in the target aspect probability distribution as supplementary information for content-based prediction. For evaluation, we test six state-of-the-art ALD models across seven datasets focused on different aspects and collected from different domains. Our proposed model rivals or exceeds those ALD methods on all of the evaluated datasets. The contributions of the paper can be summarised as follows: 1) We perform a rigorous comparison of six state-of-the-art ALD models across seven ALD benchmark datasets, and find those models do not embrace different types of abusive language aspects in different online communities.
2) We propose a generic new ALD algorithm that enables explicit integration of multiple aspects of abusive language, and detection of generic abusive language behaviour in different domains. The proposed model rivals state-of-the-art algorithms on ALD benchmark datasets and performs best overall.

\section{Related Work}
\subsection{ALD Datasets}
\label{section:dataset}
We briefly review the seven ALD benchmark datasets (Table \ref{tab:statistics table}), which were collected from different online community sources and focused on multiple compositions. \textbf{Waseem} \cite{waseem2016hateful} is a Twitter ALD dataset regarding the specific aspects of racist and sexist. The collected tweets were labeled into \textit{Racism}, \textit{Sexism} or \textit{None}. 
\textbf{HatEval} \cite{basile2019semeval} is a Twitter-based hate speech detection dataset released in SemEval-2019. It provides a general-level hate speech annotation, \textit{Hateful} or \textit{Non-hateful}, especially against immigrants and women. \textbf{OffEval} \cite{zampieri2019semeval} covers the Twitter-based offensive language detection task in SemEval-2019. It annotates as \textit{Offensive} or \textit{Not-offensive}, and includes insults, threats, and any form of untargeted profanity. 
\textbf{Davids} \cite{davidson2017automated} is a Twitter-based ALD dataset, which includes three classes, \textit{Hate}, \textit{Offensive} or \textit{Neither} based on the hate speech lexicon from \textit{Hatebase.org}. 
\textbf{Founta} \cite{DjouvasConstantinos2018LSCa} is a large Twitter-based ALD dataset claimed to be annotated with high accuracy based on their proposed incremental and iterative annotation method. It is annotated with four classes, \textit{Hateful}, \textit{Abusive}, \textit{Normal} or \textit{Spam}. 
\textbf{FNUC} \cite{gao2017detecting} is a hate speech detection dataset, which was collected from complete Fox News discussion threads, and annotated with the general level categories \textit{Hateful} or \textit{Non-hateful}.
\textbf{StormW}\cite{de2018hate} is a Stormfront-based hate speech detection dataset with general-level labels \textit{Hate} and \textit{NoHate}. Stormfront is a supremacist forum where people promote white nationalism and antisemitism.

\begin{table*}[t]
\centering

\begin{adjustbox}{width=.8\linewidth}
\begin{tabular}{c|c|c|c}
\hline
 \textbf{Dataset} & \textbf{Source} & \textbf{Size} & \textbf{Composition} \\ \hline
 \textbf{Waseem}\cite{waseem2016hateful} & Twitter & 16.2k & \textit{Racism(11.97\%)}, \textit{Sexism(19.43\%)},  \textit{None(68.60\%)}   \\ \hline
 \textbf{HatEval}\cite{basile2019semeval} & Twitter & 13k & \textit{Hateful(42.08\%)}, \textit{Non-hateful(57.92\%)} \\ \hline
 \textbf{OffEval}\cite{zampieri2019semeval} & Twitter & 13.2k & \textit{Offensive(33.23\%)}, \textit{Not-offensive(66.77\%)} \\ \hline
 \textbf{Davids}\cite{davidson2017automated} & Twitter & 24.8k & \textit{Hate(5.77\%)}, \textit{Offensive(77.43\%)},  \textit{Neither(16.80\%)}   \\ \hline
 \textbf{Founta}\cite{DjouvasConstantinos2018LSCa}& Twitter & 99k & \textit{Abusive(27.15\%)}, \textit{Hateful(4.97\%)}, \textit{Normal(53.85\%)}, \textit{Spam(4.97\%)} \\ \hline 
 \textbf{FNUC}\cite{gao2017detecting} & Fox News Discussion Threads & 1.5k & \textit{Hateful(28.50\%)}, \textit{Non-hateful(71.50\%)} \\ \hline
 \textbf{StormW}\cite{de2018hate} & Stormfront(forum) & 10.7k & \textit{Hate(10.93\%)}, \textit{NoHate(89.07\%)} \\ \hline
\end{tabular}
\end{adjustbox}
\setlength{\belowcaptionskip}{-10pt}
\caption{Comparison and Statistical analysis of seven benchmark datasets evaluated in this paper. The composition column represents different class aspects, and the class distribution in each dataset.}
\label{tab:statistics table}
\end{table*}

\subsection{ALD Approaches}
In the early stages, ALD was commonly addressed via hand-crafted rules and manual feature engineering. The first reported ALD work\cite{spertus1997smokey} utilised a decision tree to detect hostile messages based on heuristic rules. \newcite{yin2009detection} and \newcite{razavi2010offensive} added lexicon-based features together with semantic rules and designed a linear SVM and Naïve Bayes classifier for detecting hostile language. \newcite{djuric2015hate} first applied in ALD neural networks with the paragraph2vec \cite{le2014distributed} representation. \newcite{nobata2016abusive} introduced a Yahoo! dataset and tested it with neural networks by applying a combination of word, character-based and syntactic features. Recently, deep learning techniques have become popular in ALD. \newcite{badjatiya2017deep} tested FaxtText/Glove, Convolutional Neural Networks (CNNs), Long Short-Term Memory (LSTMs) in detecting hate speech. \newcite{park2017one} designed a HybridCNN (word-level and character-level) model on abusive tweet detection in both one-step and two-step style. Several works have applied bidirectional Gated Recurrent Unit (Bi-GRU) networks with Latent Topic Clustering (LTC) \newcite{lee2018comparative} and a a transformer-based framework \newcite{bugueno2019learning}. Some works integrated user profiling into their ALD models. \newcite{qian2018leveraging} utilised the bi-LSTM to model the historical behaviour of users to generate inter-user and intra-user representation. \newcite{mishra2018author} applied node2vec \cite{grover2016node2vec} to the constructed community graph of users to derive the user embedding. However, a user profiling-based approach is only possible when the user profiles are public and when the domain provides the user-community relation information.

\section{The MACAS ALD Model}

\begin{figure}[t]
\includegraphics[width=1\textwidth]{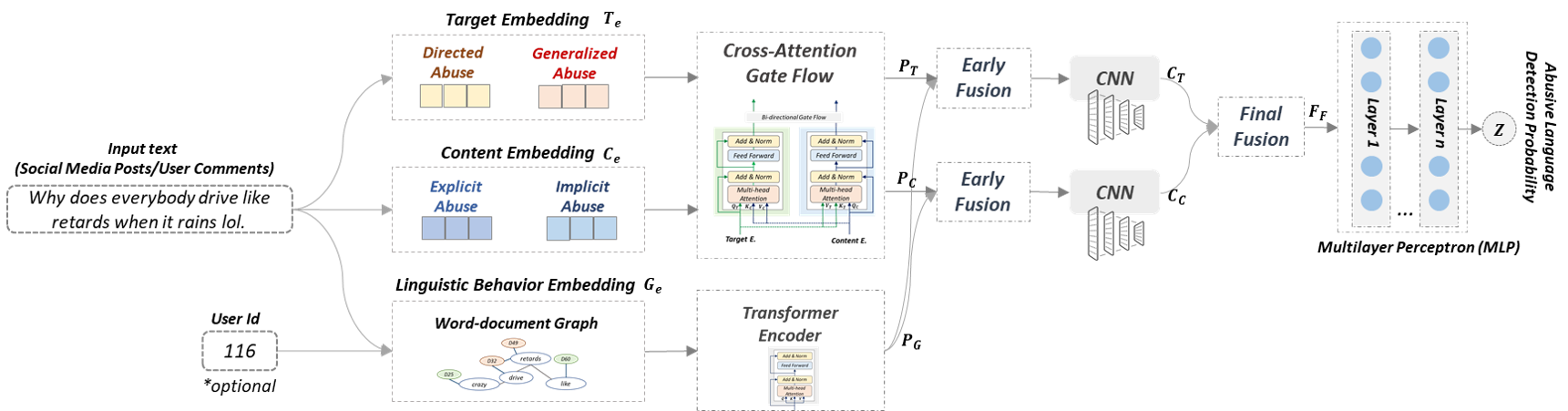}
\centering
\setlength{\belowcaptionskip}{-10pt}
\caption{The conceptual architecture of our model \textit{MACAS}}
\label{fig:architecture}
\end{figure}

We propose the \textbf{M}ulti-\textbf{A}spect \textbf{C}ross \textbf{A}ttention \textbf{S}uper Joint model for ALD. It is designed as an generic ALD that can embrace different types of abusive language aspects in different online communities. As shown in Figure \ref{fig:architecture}, MACAS can be divided into three main phases:
1) \textbf{Multi-Aspect features embedding}[Sec.\ref{MA}]. The Multi-Aspect Embedding Layer represents understanding of multi-aspects of abusive language for detecting generic abusive language behaviours. We focus on two main aspects, target and content, and each aspect has two sub-aspects. \textbf{1) Target aspect} represents  abuse directed towards either a) a specific individual/entity or b) a generalised group (e.g. gender or race). \textbf{2) Content aspect} covers a) explicit or b)implicit. Explicit abuse is unambiguous in its potential to be damaging, while implicit abusive language does not immediately impact (e.g. sarcasm). In addition to this, if the platform provides users’ historical posts, we apply Graph Convolutional Network(GCN)s to build a word-document graph embedding that represents linguistic behaviours of users. Not every online community (e.g. news forums) has user-to-user relationships (e.g. follower-following), so we avoid using user-community relationship and community network information. 
2) \textbf{Cross-Attention Gate Flow for integrating multi-aspects} [Sec.\ref{CA}] The Cross-Attention gate produces the joint integration of the target aspect and content aspect model and obtains the mutual enhancement between the two aspects. This is for producing well-integrated multi-aspects and improving the performance of generic ALD. 
3) \textbf{Final Aggregation of learned ALD embeddings}  [Sec.\ref{FF}] We aggregate multi-aspect embeddings and the user's linguistic behaviour embedding across the online post using convolutional neural networks, and produce the ALD using multi-layer-perceptron.

\subsection{Multi-Aspect Embedding Layer \footnote{In this paper, we use only four state-of-the-art natural language processing techniques that represent each abusive language aspect well. However, we expect that more techniques for each aspect embedding would produce better performance.}}
\label{MA}
\subsubsection{Target: Directed Abuse Embedding}
Directed abuse is abuse towards a specific individual or entity \cite{waseem2017understanding}. To model this aspect, a named entity recognition (NER) approach is used. To train the NER model, we apply stacked bi-directional LSTMs, which are one of the state-of-the-art models \cite{chiu2016named}. We extract the vector before the final $Softmax$ layer of the NER model and use it as the Directed Abuse Embedding.

\subsubsection{Target: Generalised Abuse Embedding}
Generalised abuse tends to target people belonging to a small set of categories, primarily gender. The gender debiasing embedding \cite{Kaneko:ACL:2019} is applied. The vocabulary set ($V$) is split into 4 mutually exclusive sets of words, namely, masculine ($V_m$), feminine ($V_f$), neutral ($V_n$) and stereotypical ($V_s$). Each word is represented by a vector which is calculated by minimising a loss function to satisfy the criteria: 1) protect the feminine information for words in $V_f$; 2) protect the masculine information for words in $V_m$; 3) protect the neutrality for words in $V_n$ (iv) remove gender biases for words in $V_s$.

\subsubsection{Content: Explicit Abuse Embedding}
For the explicit abuse, whether the target is directed or generalised, explicit abuse is usually indicated by specific keywords from the homophobic slurs lexicon. We used dict2vec \cite{tissier2017dict2vec}, which aims to learn word embeddings based on natural language dictionaries. In this paper, the model is trained by Cambridge, Collins, Oxford, dictionary.com, and we add an abusive language lexicon\footnote{http://www.rsdb.org}. 
This approach first defines strong pairs and weak pairs of words. If both words appear in each other's definition, the word pair is defined as a strong pair. If only one word appears in the other's definition, the word pair is defined as a weak pair. If the words do not appear in each other's definition they are not related. Each word is represented by a vector. Strongly paired words have more similar vectors then weakly paired words which in turn have more similar vectors than unrelated words.

\subsubsection{Content: Implicit Abuse Embedding}
Implicit abusive language does not immediately imply or denote abuse, similar to sarcasm. Here we use a hybrid of CNN and LSTM-based sarcasm detection models \cite{ghosh2016fracking}. The vector before the final $Softmax$ layer of the sarcasm detection model is the Implicit Abuse Embedding.

\subsubsection{Additional: User Linguistic Behaviour Embedding}
We model the graph by setting each comment in the training set as a document. The vocabulary is the set of all words in the documents. The corpus is the collection of all documents. The nodes of our graph are the union of the documents and the vocabulary. An edge weighted 1 exists between each node and itself. An edge exists between a document and a word if the word is in that document. The edge is weighted with the TF-IDF for the (document, word) pair, within the corpus. An edge exists between two words if they have a non-negative point-wise mutual information (PMI) with a sliding window size of 20, within the corpus. The weight for the edge is the PMI for the word pair. The edge weightings are compiled into an adjacency matrix combined with the graph's degree matrix and passed into a 2 layer GCN trained to map each document to each user as a label. For datasets without user id provided, we use the actual classification target as the document node label. From this network, we obtain embeddings for each node, that is an embedding of each document or each word. The trained word embeddings $G_{e}$ are fed into transformer encoders to get linguistic behaviour outputs.

\subsection{Cross-Attention Gate Flow}
\label{CA}

In the Cross-Attention Gate Flow, first, we use a cross transformer encoder for refining our four types of embedding: Directed abuse embedding $D$, Generalized abuse embedding $G$, Explicit abuse embedding $E$ and Implicit abuse embedding $I$. Before putting them into the cross transformer encoders, we combine $D$ with $G$ as Target embedding $T_{e}$ and broadcast $I$ to sequence length $N$, them combine it with $E$ as Content embedding $C_{e}$. Normally, for the transformer encoder \cite{vaswani2017attention}, the attention is calculated using key ($K$ of dimension $d_{k}$), query ($Q$), value ($V$):
\begin{gather}
    Attention(Q,K,V) = softmax(\frac{QK^T}{\sqrt{d_k}})V. 
\end{gather}

\begin{figure}[t]
\centering
     \begin{subfigure}{0.45\textwidth}
     \centering
         \includegraphics[width=0.45\linewidth]{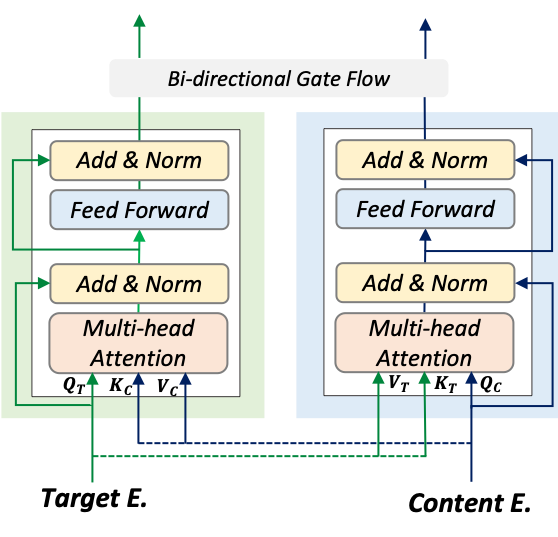}
         \caption{\textbf{C}AGF at the \textbf{B}eginning}
         \label{fig:CB}
     \end{subfigure} 
     \begin{subfigure}{0.45\textwidth}
     \centering
         \includegraphics[width=0.45\linewidth]{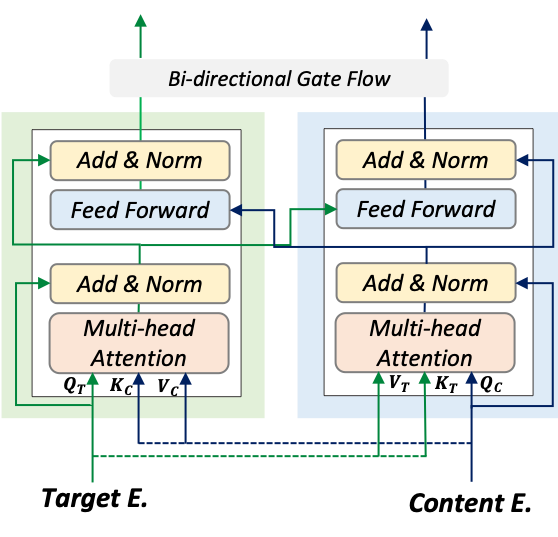}
         \caption{\textbf{C}AGF at the \textbf{B}eginning and the \textbf{M}iddle}
         \label{fig:CBM}
     \end{subfigure}
     \setlength{\belowcaptionskip}{-10pt}
     \caption{Variances of Cross-Attention Gate Flow}
     \label{fig:cross-transformer}
\end{figure}

\noindent However, to produce the joint integration of target aspect model and content aspect model, we apply the cross-transformer to $T_{e}$ and $C_{e}$. As shown in the Figure \ref{fig:cross-transformer} for each transformer encoder, we have K,Q,V for  $T_{e}$ and $C_{e}$. The K,V of $T_{e}$ and $C_{e}$ are switched, which means K,V of $T_{e}$ goes to the transformer encoder of $C_{e}$ and K,V of $C_{e}$ goes to $T_{e}$’s encoder. Then attention is calculated by
% \begin{tabularx}{\textwidth}{@{}XX@{}}
%   \begin{equation}
%   Attention_{content} = softmax(\frac{Q_cK_t^T}{\sqrt{d_k}})V_t
%   \end{equation} &
%   \begin{equation}
%   Attention_{target} = softmax(\frac{Q_tK_c^T}{\sqrt{d_k}})V_c  
%   \end{equation} 
% \end{tabularx}
\begin{equation}
    Attention_{content} = softmax(\frac{Q_cK_t^T}{\sqrt{d_k}})V_t, \quad
    Attention_{target} = softmax(\frac{Q_tK_c^T}{\sqrt{d_k}})V_c  
\end{equation}

We call the cross transformer here \textit{Cross at Beginning(CB)}. Similar to the original transformer encoder, each encoder contains one or more encoder stack(s), which mainly consists of two sub-layers: a multi-head attention layer and a fully connected feed-forward neural network (FNN). A residual connection followed by layer normalization is employed around each of the two sub-layers before feeding to the next sub-layer.
Another way to produce the joint integration occurs before the FNN layer. The output of Multi-Head Attention will be the input for the FNN layer, and then an Add \& Norm layer is applied. Normally, the output of transformer encoder is calculated by
\begin{gather}
     Output = norm(FNN(O_{MHA})+O_{MHA})
\end{gather}
The input for FNN can also be switched for Content and Target, which is called \textit{Cross in the Middle (CM)}, the output of transformer encoder will be calculated by
\begin{equation}
  T_{h} = norm(FNN(C_{MHA})+T_{MHA}), \quad C_{h} = norm(FNN(T_{MHA})+C_{MHA})
\end{equation} 

% \begin{gather}
%     T_{h} = norm(FNN(C_{MHA})+T_{MHA})  \\
%     C_{h} = norm(FNN(T_{MHA})+C_{MHA})
% \end{gather}
If the cross happens both at the beginning and in the middle, the structure will be called \textit{Cross at the Beginning and in the Middle (CBM)}. The comparison of different cross transformer structures will be discussed in \ref{ATestIntegration}. 
Both of the input embeddings $T_{e}$ and $C_{e}$ are of shape [$N$, $D_{e}$], where $D_{e}$ is the sum of the dimension of the concatenated embedding. The transformer encoder will output $T_{h}$ and $C_{h}$ in the same shape [$N$, $D_{e}$]. The hidden state of encoders $T_{h}$ from $T_{e}$ and $C_{h}$ from $C_{e}$ will be used to compute the initial abusive language probability, which is the major input of our bi-directional aspect gate flow.

%The inputs are $T_{h}$ and $C_{h}$ from the previous layer encoder. The content gate computes the abusive language probability $T_{p}$ based on the encoder target state $T_{h}$ and fuses $T_{p}$ with encoder content state $C_{h}$, resulting in the augmented content state $O_{C}$. Conversely, the target gate calculates the abusive language probability $C_{p}$ based on the encoder content state $C_{h}$ and combined $C_{p}$ with encoder target state $T_{h}$, generating the augmented target state $O_{T}$.
On top of the Cross-Attention, we introduce the Bi-directional Aspect Gate Flow that contains two gates: content gate and target gate. Denote the input sequences to our gates from the previous layer encoder as $T_h \in \mathbb{R}^{N\times D_T}$ and $C_h \in \mathbb{R}^{N\times D_C}$ where $N$ is the sequence length while $D_{T}$ and $D_{C}$ equal to dimension of target embedding and content embedding respectively. In the content gate, we first flatten $T_h$ to be $T_{hf} \in \mathbb{R}^{1\times (N*D_T)}$. We then pass $T_{hf}$ through a dense layer and apply the $Softmax$ function. The resultant $P_{Th}$ is a $D$-dimensional probability vector, where $D=N_{cls}$ is the number of distinct labels to classify, $W_C\in \mathbb{R}^{D\times D_C}$ is the weight matrix and $b_C\in \mathbb{R}^{1\times D}$ is the bias vector. Then we broadcast $P_{Th}$ over $N$ tokens. This yields $\hat{P_{Th}}\in \mathbb{R}^{N\times D}$. Then we concatenate $\hat{P_{Th}}$ with transformer encoder output state $C_{h}$ from content source, generating the augmented content state $O_C\in \mathbb{R}^{N\times (D+D_C)}$. We then again flatten $O_C$ and pass the output to the dense layer, producing an output matrix $P_C \in \mathbb{R}^{1\times D}$.
% \begin{gather}
%     P_{Th} = softmax \left(T_{hf}\cdot W_C+b_C\right) \label{eq:p_l}\\
%     \hat{P_{Th}}= repeat(P_{Th}), N \ \text{times} \label{eq:p_lrepeat}\\
%     O_C = \hat{P_{Th}} \oplus C_h \label{eq:s_l}\\
%     P_C=O_C\cdot W_C'+b_C' \label{eq:p_lo}
% \end{gather}

The procedure in the target gate is almost the same as the content gate. Here we flattened the input sequence $C_h$, generating the flattened output $C_{hf} \in \mathbb{R}^{1\times (N*D_C)}$. We then pass the result through a dense layer and apply the $Softmax$ function. The resultant $P_{Ch}$ is also broadcast to be $\hat{P_{Ch}}$ and then concatenated with the target encoder output state $T_h$, where $O_T\in \mathbb{R}^{N\times (D+D_T)}$ is the augmented target state as output matrix. Finally, $O_T$ is also flattened and then passed to the dense layer, which produces the output matrix $P_T \in \mathbb{R}^{1\times D}$.
% \begin{gather}
%     P_{Ch}=softmax \left(C_{hf}\cdot W_T+b_T\right) \label{eq:p_t}\\
%     \hat{P_{Ch}}= repeat(P_{Ch}), N \  \text{times} \label{eq:p_trepeat}\\
%     O_T = \hat{P_{Ch}} \oplus T_h \label{eq:s_t}\\
%     P_T=O_T\cdot W_T'+b_T' \label{eq:p_to}
% \end{gather}

\subsection{Final Fusion}
\label{FF}
We propose a hierarchical fusion, which fuses linguistic behaviour outputs ($P_{G}$) with content gate output ($P_{C}$) and target gate output ($P_{T}$) respectively and uses two CNNs to integrate that fusion to get $C_{C}$ and $C_{T}$, then we concatenate $C_{C}$ and $C_{T}$ then flatten it to $F_{F}$. Finally, a multi-layer perceptron (MLP) is used for final prediction: 
\begin{equation}
  L_1 = ReLU(W_1 \cdot F_{F} + b_1), \quad L_2 = ReLU(W_2 \cdot L_1 + b_2), \quad Z = softmax(W_3 \cdot L_2 + b_3)
\end{equation}
Three layers are stacked. For the each layer, $W_i$ and $b_i$ represent the weight matrix and bias vector, and the ReLU activation function is used for the first two layers. For the last layer, to get the probability of each class $Z$, softmax layer is used.

\section{Evaluation Setting}
We conducted experiments on all seven datasets with and without GCN as well as using the three different types of cross-transformer variances, which will be discussed in \ref{ATestIntegration}. The GCN embedding dimension for this linguistic behaviour graph is $D_{LBG}=200$. For transformer encoder configuration, we used dropout rate = 0.5, encoder number = 2, head number = 3, and hidden dimension = 1296. The models are trained with batch size = 16, and lr(learning rate) and number of epochs differ: \textbf{Waseem}: lr = 4e-4, epochs = 6, \textbf{HatEval}: lr = 1e-7, epochs = 6, \textbf{OffEval}: lr = 1e-7, epochs = 13, \textbf{Davids}: lr = 4e-4, epochs = 6, \textbf{Founta}: lr = 1e-5, epochs = 8, \textbf{FNUC}: lr = 1e-6, epochs = 13, \textbf{StormW}: lr = 1e-6, epochs = 7. The hyper-parameters are decided by splitting the training set into 90:10 training and validation set.

The followings are the models evaluated in our experiments. \textbf{TF-IDF features and SVM Classifier (TIS)}: TIS \cite{yin2009detection} applies TF-IDF with SVM Classifier to detect abusive language. First, TF-IDF weights of words are generated and a Support Vector Machine with radial basis function (RBF) kernel is trained to classify different kinds of abusive languages. \textbf{One-Two Steps Hybrid CNN (OTH)}: OTH \cite{park2017one} used a Hybrid CNN (word-level and character-level) model and applied it to abusive tweet detection. We applied Chars2vec as a character embedding and Glove as a word embedding. The convolutional layers with kernel size 256, 128, and 64 are stacked, and the model is trained using learning rate 4e-5 with 10 epochs. \textbf{Multi-Features with RNN (MFR)}: MFR \cite{mehdad2016characters} used a hybrid character-based and word-based Recurrent Neural Network (RNN) model to detect abusive language. After the Chars2vec and Glove embeddings, there is a vanilla stacked RNN. Three RNN layers with hidden dimensions 128, 128, and 64 are stacked, and the model is trained using learning rate 4e-6 with 10 epochs. \textbf{Two-step Word-level LSTM (TWL)}: TWL \cite{badjatiya2017deep} produced LSTM-derived representations with a Gradient Boosted Decision Trees classifier. The model applied LSTM to Glove embeddings, and the results are fed into the model. Three LSTM layers with hidden dimensions 128,128,64 are stacked, and the model is trained using learning rate 4e-6 with 10 epochs. \textbf{Latent Topic Clustering with Bi-GRU (LTC)}: LTC \cite{lee2018comparative} applies a Bi-GRU with latent topic clustering, which extracts the topic information from the aggregated hidden states of the two directions of the Bi-GRU. Three Bi-GRU layers with hidden dimensions 128, 128, and 64 are stacked, and the model is trained using learning rate 4e-5 with 10 epochs. \textbf{Character-based Transformer (CBT)}: CBT \cite{bugueno2019learning} uses a transformer-based classifier with Chars2vec embeddings. Transformer encoders with hidden dimension 400, learning rate 4e-6 with 3 epochs are used.

\section{Experiments and Results}
\subsection{Performance Comparison}

In this part, we compare our model with six baseline models over all seven datasets, discussed in Sec \ref{section:dataset}. These baseline models are constructed with various word representations as well as different neural networks or classifiers. Table \ref{evaluation1table} presents the weighted average f1 performance of each baseline model and our model over each dataset. Our model outperforms the baseline models for all these seven datasets. Applying multiple aspect embeddings enables our model to process the texts from multi-perspective views. The  Cross-Attention gate flow makes it possible to obtain the mutual enhancement between the two different aspects. Although some of the baseline models such as OTH, MFR also combine two embedding approaches (Chars2vec and Glove) to get more information, they still just consider the general information of the texts rather than extract information in a targeted fashion from various aspects. For these reasons our model can achieve performance above the baseline models.

As well as comparing our model with the baseline models, we also make some observations from comparing the six baseline models amongst themselves. Firstly, OTH and MFR use the combined embeddings of Chars2vec and Glove which gives more information. So, they can achieve relatively better weighted average f1 scores compared to most other baseline models which just use a single embedding method. Secondly, the results of TWL and LTC indicate that the bi-directional recurrent neural network leads to better performance than the simple forward recurrent neural network. This means that not only the future states but also the past ones will affect the prediction results. Thirdly, although we may not consider TF-IDF with SVM to be as good as Chars2vec or Glove with deep neural networks, TIS baseline model never gets the worst weighted f1 score for the seven datasets when compared with other models. In fact it even outperforms other baseline models on \textbf{Waseem} and \textbf{Founta}. For both datasets, there might be some particular words which are really significant for identifying the class. So TF-IDF can achieve good results for these two datasets.

\begin{table*}[t]
\fontsize{6}{7.2}\selectfont
\centering

\begin{adjustbox}{width=.7\textwidth}
\begin{tabular}{c|c|c|c|c|c|c|c}
\hline
\textbf{Dataset/Algorithm}  & \textbf{TIS}  & \textbf{OTH}  & \textbf{MFR}  & \textbf{TWL}  & \textbf{LTC}  & \textbf{CBT}  & \textbf{Ours}\\ \hline
Waseem\cite{waseem2016hateful} & \cellcolor{green!30}83.56 & 79.10 & 62.39 & 73.88 & 79.94 & 79.11 & \cellcolor{green}86.00\\
HatEval\cite{basile2019semeval} & 41.63 & 40.48 & \cellcolor{green!30}53.17 & 52.03 & 53.14 & 49.25 & \cellcolor{green}53.97\\
OffEval\cite{zampieri2019semeval} & 75.37 & 76.84 & 55.59 & 67.15 & \cellcolor{green!30}77.90 & 58.71 & \cellcolor{green}78.80\\
Davids\cite{davidson2017automated} & 88.11 & 88.37 & 79.44 & 83.74 & 87.56 & \cellcolor{green!30}88.94 & \cellcolor{green}90.34\\
Founta\cite{DjouvasConstantinos2018LSCa} & \cellcolor{green!30}79.58 & 78.59 & 73.64 & 75.23 & 79.49 & 72.04 & \cellcolor{green}80.36\\
FNUC\cite{gao2017detecting} & 68.92 & 64.51 & \cellcolor{green!30}70.71 & 65.67 & 69.78 & 67.07 & \cellcolor{green}73.20\\
StormW\cite{de2018hate} & 82.73 & \cellcolor{green!30}85.48 & 82.06 & 81.91 & 83.83 & 82.90 & \cellcolor{green}85.86\\
\hline
\end{tabular}
\end{adjustbox}
\setlength{\belowcaptionskip}{-10pt}
\caption{Overall f1 results from seven ALD models (including \textit{MACAS}) evaluated across all seven benchmark datasets. We highlight the top 2 models for each dataset, using darker colors for higher performance. For all the benchmark datasets, we train models on the train split and report results on test splits.}
\label{evaluation1table}
\end{table*}

\begin{table*}[t]
\fontsize{7}{8.4}\selectfont

\centering
\begin{tabularx}{\textwidth}{c|b|c|c|s}
\hline
\textbf{Dataset}  & \textbf{Abusive Example}  & \textbf{Actual C.}  & \textbf{Correctly C.} & \textbf{Incorrectly C.}\\ \hline
Waseem & Really \textcolor{blue}{\textbf{bitch}} really? Get on the runway and \textcolor{blue}{\textbf{back in the kitchen}}.
& Sexism & TIS, Ours & OTH, MFR, TWL, LTC, CBT
\\ \hline
Waseem & thegeek\_\textcolor{blue}{\textbf{chick}} Please answer... & Sexism & TIS, Ours & OTH, MFR, TWL, LTC, CBT \\ \hline

Waseem & RT randomfox: \#ManSpreading \textcolor{blue}{\textbf{\#FeminismIsAwful \#WomenAgainstFeminism}} & Sexism & TIS, Ours & OTH, MFR, TWL, LTC, CBT \\ \hline

Waseem & RT @asredasmyhair: \textcolor{blue}{\textbf{Fems, for the love of God}}, please stop propagating the wage gap myth. \#FemFreeFriday & Sexism & TIS, Ours & OTH, MFR, TWL, LTC, CBT \\ \hline

FNUC & \textcolor{blue}{\textbf{BLM is a Hate group}} Given Birth by Obama's comments That put a target on the backs of police officers 
& Hateful & Ours & TIS, OTH, MFR, TWL, LTC, CBT
\\ \hline
StormW & i am a \textcolor{blue}{\textbf{white nationalist}} of a christian faith but still am a \textcolor{blue}{\textbf{white nationalist}} for racial survival the anti racist christians are the true chirstians enemy & Hate & OTH, Ours & TIS, MFR, TWL, LTC, CBT
\\ \hline
\end{tabularx}
\setlength{\belowcaptionskip}{-10pt}
\caption{Some prediction example from our model \textit{MACAS} compared to the baselines. Actual C.: Actual Class, Correctly C.: Correctly Classified Models, Incorrectly C.: Incorrectly Classified Models.}
\label{evaluation1table2}
\end{table*}

Next, an analysis of examples to show how different models perform on different datasets is conducted, and the results are shown in Table \ref{evaluation1table2}. The first example is from \textbf{Waseem}, “Really bitch really? Get on the runway and back in the kitchen.”, which should be predicted as Sexism. It is quite explicit in that the word “bitch” is in this sentence, and this makes TIS predict it as Sexism easily since TF-IDF is focusing on the word occurrence. Besides, “back in the kitchen” is implicit Sexism, implying women should be in the kitchen. The similar patterns can be found from the second instance “thegeek\_chick please answer” by explicitly mentioning the word ‘chick’. The third and fourth samples represent abusive language or hate speech about the topic Feminism. The third explicitly stated the words ‘Feminism’ and ‘Awful’ and TIS and our model successfully detected the abuse with an explicit hate speech aspect identification. Our model, which considers the explicit and implicit aspects, can predict the sentence as Sexism easily. Another example is from \textbf{FNUC}, “BLM is a Hate group Given Birth by Obama's comments That put a target on the backs of police officers” which should be Hateful. This comment insults the “Black Life Matters” by calling it a Hate Group. Normally, describing something as a hate group is not hate speech, but in this case, calling BLM a hate group is racism. This is not easy for the baseline models to spot, and only our model predicts it correctly. For the last example from \textbf{StormW}, “i am a white nationalist of a christian faith but still am a white nationalist for racial survival the anti racist christians are the true chirstians enemy”, the user described himself as “white nationalist” which is one kind of hate speech, and OTH can predict this sentence as Hate. The reason is that the CNN used in OTH can capture the information for phrases, which is the “white nationalist” here. Besides, our model can predict this sentence correctly since the sentence is a general explicit hate speech.

\subsection{Ablation Testing - Cross-attention gate flow}
\label{ATestIntegration}

In this part, three different structures of cross transformer encoders are tested: 1) Cross-transformer at the beginning of the the transformer encoder \textbf{(CB)}: exchanging content's and target's K and V at the beginning of the transformer encoders as in Figure \ref{fig:cross-transformer}; 2) Cross-transformer in the middle of the transformer encoder \textbf{(CM)}: exchanging content's and target’s input for Feed Forward layer in the transformer encoder, which is in the middle of the transformer encoders; 3) Cross-transformer at both places \textbf{(CBM)}: the combination of CB and CM. Due to the poor performance of CM, only results for 7 datasets with CB and CBM structure are shown in Table \ref{evaluation2table}. Besides, to find whether and how GCN is improving the performance of our model, different structures are also compared: 1) Model without GCN; 2) Model with GCN using hierarchical fusion, repeating one or three times. We show one and three times here because on all the datasets our model achieves the best performance with one or three repeated fusions when GCN is also used. Two conclusions are drawn based on the results of CB and CBM:

\begin{table*}[t]
\fontsize{6}{7.2}\selectfont
\centering

\begin{adjustbox}{width=.7\textwidth}
\begin{tabular}{c|c|c|c|c|c|c|c}
\hline
\textbf{Methods}  & \textbf{Waseem}  & \textbf{HatEval}  & \textbf{OffEval}  & \textbf{Davids}  & \textbf{Founta}  & \textbf{FNUC}  & \textbf{StormW}\\ \hline
CB, no G & 82.35 & \cellcolor{green}53.97 & \cellcolor{green}78.80 & \cellcolor{green}90.34 & \cellcolor{green}80.36 & 65.31 & 82.90\\ 
CB, G, N=1 & \cellcolor{green}86.00 & 51.88 & 75.06 & 87.36 & 76.90 & \cellcolor{green}73.20 & 84.52\\
CB, G, N=3 & 83.76 & 42.71 & 75.03 & 88.24 & 75.42 & 68.39 & \cellcolor{green}85.86\\
CBM, no G & 81.53 & \cellcolor{green!30}53.28 & \cellcolor{green!30}77.37 & \cellcolor{green!30}90.25 & \cellcolor{green!30}80.28 & 65.67 & 84.14\\
CBM, G, N=1 & \cellcolor{green!30}85.22 & 39.91 & 72.60 & 90.12 & 76.06 & \cellcolor{green!30}68.92 & 85.09\\
CBM, G, N=3 & 82.77 & 42.86 & 75.10 & 90.11 & 77.03 & 68.16 & \cellcolor{green!30}85.12\\ \hline
\end{tabular}
\end{adjustbox}
\setlength{\belowcaptionskip}{-10pt}
\caption{Abusive language detection results across seven benchmark datasets for \textit{MACAS} with two cross attention aspect gate flow mechanisms and graph embedding. We highlight the top 2 settings for each dataset. The darker the colour, the better the performance. The comparison provides different parameters (\textit{N}) of final fusion layers, including N=1 or 3. (CB: cross-attention at the beginning, CBM: cross-attention at the beginning and the middle, G: the user linguistic behaviour graph embedding)}
\label{evaluation2table}
\end{table*}

Firstly, the best model is always the CB model, and the second best is always the CBM model with the same GCN structure. So comparing between CB and CBM structure, CB has a better performance and we use this structure as our final model. Besides, in most cases, CB outperforms CBM if they share the same GCN structure, which also shows that, overall, CBM is worse than CB. Considering the fact that CM is the worst, we can say that cross in the middle transformer encoder will lower the model performance. Exchanging content’s and target’s K,V is important since it allows target aspects to query on the content aspects and vice versa. However, exchanging values before Feed Forward Layer only gives a different add and norm which doesn’t increase the interaction between content aspects and target aspects usefully.

Secondly, our model can have a better performance with GCN when there is user id in the dataset. Not all the datasets provide user id, and as mentioned in Sec \ref{MA}, User Linguistic Behavior embedding is trained by using the user id as the target. For those datasets without userid, the real abusive labels are used as the training target. By comparison, we can find that \textbf{Waseem}, \textbf{StormW}, and \textbf{FNUC} which provide user id in the datasets have a better performance using a model with GCN, and the other four datasets, which don’t provide user id, have a better performance using a model without GCN. Therefore, for the dataset with user id, User Linguistic Behavior which is from GCN, can improve the performance of our model. And for those datasets without user id, the model structure without GCN is recommended.

\subsection{Ablation Testing - Multi-aspect embedding}

\begin{table*}[t]
\fontsize{6}{7.2}\selectfont
\centering

\begin{adjustbox}{width=.7\textwidth}
\begin{tabular}{c|c|c|c|c|c|c|c}
\hline
\textbf{Combinations} & \textbf{Waseem} & \textbf{HatEval} & \textbf{OffEval} & \textbf{Davids} & \textbf{Founta} & \textbf{FNUC} & \textbf{StormW}\\
\hline
D + E & 80.16 & 49.94 & 75.81 & 89.58 & 80.02 & 66.03 & 82.23\\
D + I & \cellcolor{red!25}61.93 & \cellcolor{red!25}47.04 & \cellcolor{red!25}54.63 & \cellcolor{red!25}68.27 & \cellcolor{red!25}67.88 & 64.51 & \cellcolor{red!25}81.91\\
D + E + I & 80.57 & 47.11 & 69.95 & 87.11 & 79.80 & \cellcolor{red!25}64.03 & \cellcolor{red!25}81.91\\
G + E & 79.67 & 52.78 & 76.95 & 86.92 & 79.39 & 65.56 & \cellcolor{green!30}84.85\\
G + I & 80.10 & 53.63 & 57.38 & 87.52 & 79.35 & 64.24 & \cellcolor{red!25}81.91\\
G + E + I & 79.12 & 48.71 & 72.17 & 89.19 & 79.14 & 65.96 & 82.04\\
D + G + E & 78.63 & 53.60 & 73.78 & 88.51 & 80.23 & \cellcolor{green}\textbf{68.28} & 82.44\\
D + G + I & 79.74 & 52.65 & 75.12 & 89.76 & 79.76 & 65.31 & 83.44\\
D + G + E + I & \cellcolor{green}\textbf{82.35} & \cellcolor{green}\textbf{53.97} & \cellcolor{green}\textbf{78.80} & \cellcolor{green}\textbf{90.34} & \cellcolor{green}\textbf{81.57} & 65.31 & \cellcolor{green}\textbf{83.93}\\ \hline
\end{tabular}
\end{adjustbox}
\setlength{\belowcaptionskip}{-10pt}
\caption{Ablation studies comparing different types integration of multi-aspects for the generic ALD model. In the proposed model, \textit{MACAS}, we introduced four aspect embeddings, including directed abuse (D), generalised abuse (G), explicit abuse (E), and implicit abuse (I). Directed and generalised abuses are in the group of a target aspect, while explicit and implicit abuses are in a content aspect group. The ablation testing is conducted in a different combination of aspect embedding from each higher-level of aspect groups. The highest performance is highlighted in green, the lowest is marked in red.}
\label{evaluation3table}
\end{table*}

To check how aspect embeddings contribute to the model, an ablation test on different combinations of the embeddings is conducted on all these seven datasets. We use the CB model without GCN for the prediction. Table \ref{evaluation3table} presents the weighted average f1 scores for 9 different combinations of four aspect embedding models, including Directed abuse $D$, Generalised abuse $G$, Explicit abuse $E$, and Implicit abuse $I$. Each target and content aspect should include at least one embedding. 

For \textbf{Waseem}, the $D+G+E+I$ combination achieves the best performance with the weighted average f1 score 82.35 and most other combinations have a slightly lower performance. In contrast, $D+I$ gets the worst weighted f1 score of 61.93. The reason why $D+I$ is much worse than other combinations may lie in two facts: 1) In this dataset, abusive language is generally more explicit rather than directly aiming at a specific target in an implicit way. 2) Even humans can not distinguish Direct Abuse in an Implicit way easily, and it can be very difficult for the annotators to annotate the label correctly. Besides, the $D+G+E+I$ combination outperforms other cases because it takes all the aspects into consideration. Similar results occur on other Twitter datasets \textbf{Davids}, \textbf{HatEval}, \textbf{OffEval} and \textbf{Founta}, $D+G+E+I$ achieves the best while $D+I$ is much worse. For \textbf{FNUC}, due to the small volume of dataset and imbalanced labels, not all the combinations have a good prediction result. $D+G+E$ having the best performance implies that the dataset doesn’t have a large number of implicit abuse samples. For \textbf{StormW}, $D+G+E+I$ gets the best performance. Besides, $G+E$ also has a good performance. The reason is that this dataset is collected from a racism forum and most hate speech on that website is generally abusive in an explicit way. Based on the analysis of the different embedding combinations on these datasets, we can conclude that the embeddings used may vary based on different kinds of datasets, but combining them all is always a good idea. 
Although four specific different embeddings are selected in our model to represent four different aspects, other kinds of embeddings could also be used as long as they can represent the corresponding aspects.

\section{Conclusion}
Abusive language detection is an essential but challenging task, and it is almost impossible to successfully encompass all different abusive language tasks in different domains. The evaluation also shows that most of the state-of-the-art ALD algorithms do not generalise their model to different types of abusive language problems or datasets. In this paper, we proposed a new generic abusive language model, called MACAS, which applied multi-aspect embeddings to represent generalised characteristics of the domain and introduced a cross-attention gate flow model to achieve better performance by mutual enhancement between the target aspect and the content aspect. The results indicate that our framework was successful and effective in capturing abusive language aspects in different domains. Compared to other ALD models, our model successfully works in general abusive language detection, and it is hoped that MACAS provides some insight into the future direction of generic abusive language detection.


\bibliographystyle{coling}
\bibliography{coling2020}

\begin{thebibliography}{}

\bibitem[\protect\citename{Badjatiya \bgroup et al.\egroup
  }2017]{badjatiya2017deep}
Pinkesh Badjatiya, Shashank Gupta, Manish Gupta, and Vasudeva Varma.
\newblock 2017.
\newblock Deep learning for hate speech detection in tweets.
\newblock In {\em Proceedings of the 26th International Conference on World
  Wide Web Companion}, pages 759--760. International World Wide Web Conferences
  Steering Committee.

\bibitem[\protect\citename{Basile \bgroup et al.\egroup
  }2019]{basile2019semeval}
Valerio Basile, Cristina Bosco, Elisabetta Fersini, Debora Nozza, Viviana
  Patti, Francisco Manuel~Rangel Pardo, Paolo Rosso, and Manuela Sanguinetti.
\newblock 2019.
\newblock Semeval-2019 task 5: Multilingual detection of hate speech against
  immigrants and women in twitter.
\newblock In {\em Proceedings of the 13th International Workshop on Semantic
  Evaluation}, pages 54--63.

\bibitem[\protect\citename{Bugue{\~n}o and Mendoza}2019]{bugueno2019learning}
Margarita Bugue{\~n}o and Marcelo Mendoza.
\newblock 2019.
\newblock Learning to detect online harassment on twitter with the transformer.
\newblock In {\em Joint European Conference on Machine Learning and Knowledge
  Discovery in Databases}, pages 298--306. Springer.

\bibitem[\protect\citename{Chiu and Nichols}2016]{chiu2016named}
Jason~PC Chiu and Eric Nichols.
\newblock 2016.
\newblock Named entity recognition with bidirectional lstm-cnns.
\newblock {\em Transactions of the Association for Computational Linguistics},
  4:357--370.

\bibitem[\protect\citename{Davidson \bgroup et al.\egroup
  }2017]{davidson2017automated}
Thomas Davidson, Dana Warmsley, Michael Macy, and Ingmar Weber.
\newblock 2017.
\newblock Automated hate speech detection and the problem of offensive
  language.
\newblock In {\em Eleventh international aaai conference on web and social
  media}.

\bibitem[\protect\citename{de Gibert \bgroup et al.\egroup }2018]{de2018hate}
Ona de~Gibert, Naiara Perez, Aitor Garc{\'\i}a-Pablos, and Montse Cuadros.
\newblock 2018.
\newblock Hate speech dataset from a white supremacy forum.
\newblock In {\em Proceedings of the 2nd Workshop on Abusive Language Online
  (ALW2)}, pages 11--20.

\bibitem[\protect\citename{Djouvas \bgroup et al.\egroup
  }2018]{DjouvasConstantinos2018LSCa}
Constantinos Djouvas, Despoina Chatzakou, Ilias Leontiadis, Jeremy Blackburn,
  Gianluca Stringhini, Athena Vakali, Michael Sirivianos, and Nicolas
  Kourtellis.
\newblock 2018.
\newblock Large scale crowdsourcing and characterization of twitter abusive
  behavior.
\newblock {\em arXiv.org}.

\bibitem[\protect\citename{Djuric \bgroup et al.\egroup }2015]{djuric2015hate}
Nemanja Djuric, Jing Zhou, Robin Morris, Mihajlo Grbovic, Vladan Radosavljevic,
  and Narayan Bhamidipati.
\newblock 2015.
\newblock Hate speech detection with comment embeddings.
\newblock In {\em Proceedings of the 24th international conference on world
  wide web}, pages 29--30. ACM.

\bibitem[\protect\citename{Gao and Huang}2017]{gao2017detecting}
Lei Gao and Ruihong Huang.
\newblock 2017.
\newblock Detecting online hate speech using context aware models.
\newblock In {\em Proceedings of the International Conference Recent Advances
  in Natural Language Processing, RANLP 2017}, pages 260--266.

\bibitem[\protect\citename{Ghosh and Veale}2016]{ghosh2016fracking}
Aniruddha Ghosh and Tony Veale.
\newblock 2016.
\newblock Fracking sarcasm using neural network.
\newblock In {\em Proceedings of the 7th workshop on computational approaches
  to subjectivity, sentiment and social media analysis}, pages 161--169.

\bibitem[\protect\citename{Grover and Leskovec}2016]{grover2016node2vec}
Aditya Grover and Jure Leskovec.
\newblock 2016.
\newblock node2vec: Scalable feature learning for networks.
\newblock In {\em Proceedings of the 22nd ACM SIGKDD international conference
  on Knowledge discovery and data mining}, pages 855--864.

\bibitem[\protect\citename{Kaneko and Bollegala}2019]{Kaneko:ACL:2019}
Masahiro Kaneko and Danushka Bollegala.
\newblock 2019.
\newblock Gender-preserving debiasing for pre-trained word embeddings.
\newblock In {\em Proc. of the 57th Annual Meeting of the Association for
  Computational Linguistics (ACL)}.

\bibitem[\protect\citename{Le and Mikolov}2014]{le2014distributed}
Quoc Le and Tomas Mikolov.
\newblock 2014.
\newblock Distributed representations of sentences and documents.
\newblock In {\em International conference on machine learning}, pages
  1188--1196.

\bibitem[\protect\citename{Lee \bgroup et al.\egroup }2018]{lee2018comparative}
Younghun Lee, Seunghyun Yoon, and Kyomin Jung.
\newblock 2018.
\newblock Comparative studies of detecting abusive language on twitter.
\newblock In {\em Proceedings of the 2nd Workshop on Abusive Language Online
  (ALW2)}, pages 101--106.

\bibitem[\protect\citename{Mehdad and Tetreault}2016]{mehdad2016characters}
Yashar Mehdad and Joel Tetreault.
\newblock 2016.
\newblock Do characters abuse more than words?
\newblock In {\em Proceedings of the 17th Annual Meeting of the Special
  Interest Group on Discourse and Dialogue}, pages 299--303.

\bibitem[\protect\citename{Metz and Issac}2019]{CadeMetz2019Facebook}
Cade Metz and Mike Issac.
\newblock 2019.
\newblock Facebook’s a.i. whiz now faces the task of cleaning it up.
  sometimes that brings him to tears.
\newblock {\em The New York Times}.

\bibitem[\protect\citename{Mishra \bgroup et al.\egroup
  }2018]{mishra2018author}
Pushkar Mishra, Marco Del~Tredici, Helen Yannakoudakis, and Ekaterina Shutova.
\newblock 2018.
\newblock Author profiling for abuse detection.
\newblock In {\em Proceedings of the 27th International Conference on
  Computational Linguistics}, pages 1088--1098.

\bibitem[\protect\citename{Nobata \bgroup et al.\egroup
  }2016]{nobata2016abusive}
Chikashi Nobata, Joel Tetreault, Achint Thomas, Yashar Mehdad, and Yi~Chang.
\newblock 2016.
\newblock Abusive language detection in online user content.
\newblock In {\em Proceedings of the 25th international conference on world
  wide web}, pages 145--153. International World Wide Web Conferences Steering
  Committee.

\bibitem[\protect\citename{Park and Fung}2017]{park2017one}
Ji~Ho Park and Pascale Fung.
\newblock 2017.
\newblock One-step and two-step classification for abusive language detection
  on twitter.
\newblock In {\em Proceedings of the First Workshop on Abusive Language
  Online}, pages 41--45.

\bibitem[\protect\citename{Qian \bgroup et al.\egroup
  }2018]{qian2018leveraging}
Jing Qian, Mai ElSherief, Elizabeth Belding, and William~Yang Wang.
\newblock 2018.
\newblock Leveraging intra-user and inter-user representation learning for
  automated hate speech detection.
\newblock In {\em Proceedings of the 2018 Conference of the North American
  Chapter of the Association for Computational Linguistics: Human Language
  Technologies, Volume 2 (Short Papers)}, pages 118--123.

\bibitem[\protect\citename{Razavi \bgroup et al.\egroup
  }2010]{razavi2010offensive}
Amir~H Razavi, Diana Inkpen, Sasha Uritsky, and Stan Matwin.
\newblock 2010.
\newblock Offensive language detection using multi-level classification.
\newblock In {\em Canadian Conference on Artificial Intelligence}, pages
  16--27. Springer.

\bibitem[\protect\citename{Spertus}1997]{spertus1997smokey}
Ellen Spertus.
\newblock 1997.
\newblock Smokey: Automatic recognition of hostile messages.
\newblock In {\em Aaai/iaai}, pages 1058--1065.

\bibitem[\protect\citename{Tissier \bgroup et al.\egroup
  }2017]{tissier2017dict2vec}
Julien Tissier, Christophe Gravier, and Amaury Habrard.
\newblock 2017.
\newblock Dict2vec : Learning word embeddings using lexical dictionaries.
\newblock In {\em Proceedings of the 2017 Conference on Empirical Methods in
  Natural Language Processing}, pages 254--263, Copenhagen, Denmark, September.
  Association for Computational Linguistics.

\bibitem[\protect\citename{Vaswani \bgroup et al.\egroup
  }2017]{vaswani2017attention}
Ashish Vaswani, Noam Shazeer, Niki Parmar, Jakob Uszkoreit, Llion Jones,
  Aidan~N Gomez, {\L}ukasz Kaiser, and Illia Polosukhin.
\newblock 2017.
\newblock Attention is all you need.
\newblock In {\em Advances in neural information processing systems}, pages
  5998--6008.

\bibitem[\protect\citename{Waseem and Hovy}2016]{waseem2016hateful}
Zeerak Waseem and Dirk Hovy.
\newblock 2016.
\newblock Hateful symbols or hateful people? predictive features for hate
  speech detection on twitter.
\newblock In {\em Proceedings of the NAACL student research workshop}, pages
  88--93.

\bibitem[\protect\citename{Waseem \bgroup et al.\egroup
  }2017]{waseem2017understanding}
Zeerak Waseem, Thomas Davidson, Dana Warmsley, and Ingmar Weber.
\newblock 2017.
\newblock Understanding abuse: A typology of abusive language detection
  subtasks.
\newblock In {\em Proceedings of the First Workshop on Abusive Language
  Online}, pages 78--84.

\bibitem[\protect\citename{Waseem}2016]{waseem2016you}
Zeerak Waseem.
\newblock 2016.
\newblock Are you a racist or am i seeing things? annotator influence on hate
  speech detection on twitter.
\newblock In {\em Proceedings of the first workshop on NLP and computational
  social science}, pages 138--142.

\bibitem[\protect\citename{Xu \bgroup et al.\egroup }2012]{xu2012learning}
Jun-Ming Xu, Kwang-Sung Jun, Xiaojin Zhu, and Amy Bellmore.
\newblock 2012.
\newblock Learning from bullying traces in social media.
\newblock In {\em Proceedings of the 2012 conference of the North American
  chapter of the association for computational linguistics: Human language
  technologies}, pages 656--666. Association for Computational Linguistics.

\bibitem[\protect\citename{Yin \bgroup et al.\egroup }2009]{yin2009detection}
Dawei Yin, Zhenzhen Xue, Liangjie Hong, Brian~D Davison, April Kontostathis,
  and Lynne Edwards.
\newblock 2009.
\newblock Detection of harassment on web 2.0.
\newblock {\em Proceedings of the Content Analysis in the WEB}, 2:1--7.

\bibitem[\protect\citename{Zampieri \bgroup et al.\egroup
  }2019]{zampieri2019semeval}
Marcos Zampieri, Shervin Malmasi, Preslav Nakov, Sara Rosenthal, Noura Farra,
  and Ritesh Kumar.
\newblock 2019.
\newblock Semeval-2019 task 6: Identifying and categorizing offensive language
  in social media (offenseval).
\newblock In {\em Proceedings of the 13th International Workshop on Semantic
  Evaluation}, pages 75--86.

\end{thebibliography}

\end{document}